\pdfoutput=1

\documentclass[11pt]{article}
\usepackage[final]{acl}

\usepackage{times}
\usepackage{latexsym}

\usepackage{amsmath}
\usepackage{amsfonts}
\usepackage{graphicx} 
\usepackage{subfigure}
\usepackage{booktabs}
\usepackage{array}
\usepackage{enumitem}
\usepackage{amsthm}
\usepackage{algpseudocode}
\usepackage[switch]{lineno}
\usepackage[utf8]{inputenc}
\usepackage{eqparbox}
\usepackage[nopar]{lipsum}
\usepackage{multirow}
\usepackage{makecell}
\usepackage{xcolor}
\usepackage{hhline} 
\usepackage{microtype}
\usepackage{diagbox}
\usepackage{adjustbox}

\usepackage{hyperref}

\usepackage{relsize}

\usepackage{booktabs,multirow,array}
\newcolumntype{N}{@{}m{0pt}@{}}

\usepackage[many]{tcolorbox}
\newtcolorbox{fancyquotes}{%
    enhanced jigsaw, 
    breakable,      
    frame hidden,   
    left=0.5cm,       
    right=0.1cm,      
    overlay={%
        \node [scale=8,
            text=black,
            inner sep=0pt,] at ([xshift=-1cm,yshift=-1cm]frame.north west){}; 
        \node [scale=8,
            text=black,
            inner sep=0pt,] at ([xshift=1cm]frame.south east){};  
            },
                parbox=false,
}

\usepackage{mathtools, nccmath}
\usepackage{scrextend}
\deffootnote[.25in]{.25in}{.15in}{\makebox[.25in][r]{\thefootnotemark .\hspace{.15in}}}

\makeatletter

\newtheorem*{proof*}{Proof}

\usepackage{algorithm}
\usepackage{listings}
\usepackage{color}

\definecolor{codeblue}{rgb}{0.25,0.5,0.5}
\def\@fnsymbol#1{\ensuremath{\ifcase#1\or \dagger\or *\or \ddagger\or
   \mathsection\or \mathparagraph\or \|\or **\or \dagger\dagger
   \or \ddagger\ddagger \else\@ctrerr\fi}}
   
\renewcommand{\arraystretch}{1.5}
\newcolumntype{C}[1]{>{\centering\let\newline\\\arraybackslash\hspace{0pt}}m{#1}}

\ExplSyntaxOn
\NewExpandableDocumentCommand { \ValuePlusOne } { m } 
  { \int_eval:n { \int_use:c { c @ #1 } + 1 } }
\NewExpandableDocumentCommand { \Sec } { m } 
  { \fp_eval:n { secd ( #1 ) } }
\NewDocumentCommand { \Rot } { m }
  { 
    \hbox_to_wd:nn { 1 em }
      { 
        \hbox_overlap_right:n 
          { 
            \skip_horizontal:n { \fp_to_dim:n { 7 * cosd (\Angle) } } 
            \rotatebox{\Angle}{#1}
          } 
      } 
  }
\ExplSyntaxOff

\def\Angle{45}
    
\bigskip
\def\Angle{90}

\title{M3-Embedding: Multi-Linguality, Multi-Functionality, Multi-Granularity Text Embeddings Through Self-Knowledge Distillation} 

\author{
Jianlv Chen$^{\clubsuit}$~
Shitao Xiao$^{\spadesuit}$\thanks{Co-first author}~~
Peitian Zhang$^{\spadesuit}$~
Kun Luo$^{\spadesuit}$~
Defu Lian$^{\clubsuit*}$~
Zheng Liu$^{\spadesuit}$\thanks{Corresponding authors}~
\\ 
$^\clubsuit$ University of Science and Technology of China \ \ \
$^\spadesuit$ BAAI \\ 
{\tt stxiao@baai.ac.cn} \ \ \
{\tt \{namespace.pt,luokun695,zhengliu1026\}@gmail.com} \\
{\tt chenjianlv@mail.ustc.edu.cn} \ \ \
{\tt liandefu@ustc.edu.cn}
}

\begin{document}
\maketitle


\begin{abstract}
In this paper, we introduce a new embedding model called \textbf{M3-Embedding}, which is distinguished for its versatility in \textit{Multi-Linguality}, \textit{Multi-Functionality}, and \textit{Multi-Granularity}. It provides a uniform support for the semantic retrieval of more than 100 working languages. It can simultaneously accomplish the three common retrieval functionalities: dense retrieval, multi-vector retrieval, and sparse retrieval. Besides, it is also capable of processing inputs of different granularities, spanning from short sentences to long documents of up to 8,192 tokens. The effective training of M3-Embedding presents a series of technical contributions. Notably, we propose a novel self-knowledge distillation approach, where the relevance scores from different retrieval functionalities can be integrated as the teacher signal to enhance the training quality. We also optimize the batching strategy, which enables a large batch size and high training throughput to improve the discriminativeness of embeddings. M3-Embedding exhibits a superior performance in our experiment, leading to new state-of-the-art results on multilingual, cross-lingual, and long-document retrieval benchmarks.\footnote{The model, code, and data is publicly available at \href{https://github.com/FlagOpen/FlagEmbedding}{https://github.com/FlagOpen/FlagEmbedding}.}


\end{abstract}

\section{Introduction}
Embedding models are a critical form of DNN application in natural language processing. They encode the textual data in the latent space, where the underlying semantics of the data can be expressed by the output embeddings \cite{reimers-gurevych-2019-sentence,ni-etal-2022-sentence}. With the advent of pre-trained language models, the quality of text embeddings have been substantially improved, making them imperative components for the information retrieval (IR) system. One common form of embedding-based IR application is dense retrieval, where relevant answers to the query can be retrieved based on the embedding similarity \cite{karpukhin-etal-2020-dense,xiong2020approximate,neelakantan2022text,wang2022text,bge_embedding}. Besides, the embedding model can also be applied to other IR tasks, such as multi-vector retrieval where the fine-grained relevance between query and document is computed based on the interaction score of multiple embeddings \cite{khattab2020colbert}, and sparse or lexical retrieval where the importance of each term is estimated by its output embedding \cite{gao-etal-2021-coil,lin2021few,dai2020context}.


\begin{figure}[t]
\centering
\includegraphics[width=1.0\linewidth]{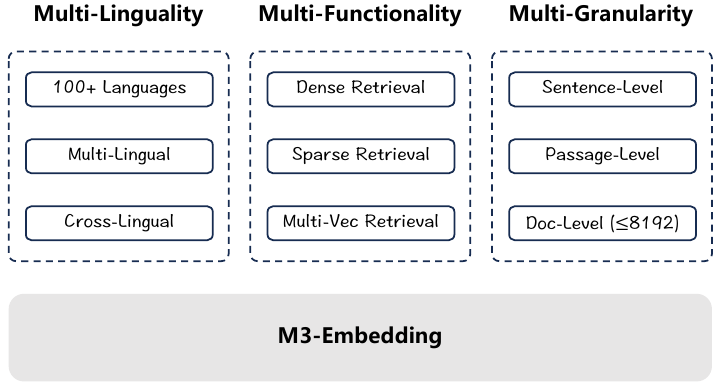}
\vspace{-15pt}
\caption{\textbf{Characters of M3-Embedding.}}
\vspace{-10pt}
\label{fig:2}
\end{figure}

Despite the widespread popularity of text embeddings, the existing methods are still limited in versatility. First of all, most of the embedding models are tailored only for English, leaving few viable options for the other languages. Secondly, the existing embedding models are usually trained for one single retrieval functionality. However, typical IR systems call for the compound workflow of multiple retrieval methods. Thirdly,
it is challenging to train a competitive long-document retriever due to the overwhelming training cost, where most of the embedding models can only support short inputs.


To address the above challenges, we introduce \textbf{M3-Embedding}, which is pronounced for its breakthrough of versatility in \textit{working languages}, \textit{retrieval functionalities}, and \textit{input granularities}. Particularly, M3-Embedding is proficient in multi-linguality, which is able to support more than 100 world languages. By learning a common semantic space for different languages, enables both multi-lingual retrieval within each language and cross-lingual retrieval between different languages. Besides, it is able to generate versatile embeddings to support different retrieval functionalities, not just dense retrieval, but also sparse retrieval and multi-vector retrieval. Finally, M3-Embedding is learned to process different input granularities, spanning from short inputs like sentences and passages, to long documents of up to 8,192 input tokens.  


The training of M3-Embedding poses a significant challenge. In our work, the following technical contributions are made to optimize the embedding quality. Firstly, we propose a novel \textbf{self knowledge distillation} framework, where the multiple retrieval functionalities can be jointly learned and mutually reinforced. In M3-Embedding, the [CLS] embedding is used for dense retrieval, while embeddings from other tokens are used for sparse retrieval and multi-vector retrieval. Based on the principle of ensemble learning \cite{buhlmann2012bagging}, such heterogenous predictors can be combined as a stronger predictor. Thus, we integrate the relevance scores from different retrieval functions as the teacher signal, which is used to enhance the learning process via knowledge distillation. Secondly, we optimize the \textbf{batching strategy} to achieve a large batch size and high training throughput, which substantially contributes to the discriminativeness of embeddings. Last but not least, we perform extensive and high-quality \textbf{data curation}. Our dataset includes three sources: 1) the extraction of unsupervised data from massive multi-lingual corpora, 2) the integration of closely related supervised data, 3) the synthesization of scarce training data. The three data sources are complement to each other and applied to different training stages, which lays a solid foundation for the versatile text embeddings.

M3-Embedding exhibits a remarkable versatility in our experiments. It achieves superior retrieval quality for a variety of languages, leading to state-of-the-art performances on popular multi-lingual and cross-lingual benchmarks like MIRACL \cite{zhang-etal-2023-miracl} and MKQA \cite{longpre-etal-2021-mkqa}. It effectively learns the three retrieval functionalities, which can not only work individually but also work together for an even stronger retrieval quality. It also well maintains its superior capability across different input granularities within 8192 tokens, which outperforms the existing methods by a notable advantage. 


Our contributions are summarized as follows. 1) We present M3-Embedding, which achieves unprecedented versatility in multi-linguality, multi-functionality, and multi-granularity. 2) We propose a novel training framework of self-knowledge distillation and optimize the batching strategy for efficient training. We also create high-quality training resource based on comprehensive data curation. 3) Our model, code, and data is publicly available, offering critical resources for both direct usage and future development of text embeddings. 


\section{Related Work}
The related works are reviewed from three aspects: general text embeddings, embedding models for neural retrieval, embeddings of multi-linguality. 

In the past few years, substantial progress has been achieved in the field of text embedding. One major driving force is the popularity of pre-trained language models, where the underlying semantic of the data can be effectively encoded by such powerful text encoders \cite{reimers-gurevych-2019-sentence,karpukhin-etal-2020-dense,ni-etal-2022-sentence}. In addition, the progress of contrastive learning is another critical factor, especially the improvement of negative sampling \cite{xiong2020approximate,qu-etal-2021-rocketqa} and the exploitation of knowledge distillation \cite{hofstatter2021efficiently,ren-etal-2021-rocketqav2,zhang2021adversarial}. On top of these well-established techniques, it becomes increasingly popular to learn versatile embedding models, which are able to uniformly support a variety of application scenarios. So far, there have been many impactful methods in the direction, like Contriever \cite{mcontriever}, LLM-Embedder \cite{llm-embedder}, E5 \cite{wang2022text}, BGE \cite{bge_embedding}, SGPT \cite{muennighoff2022sgpt}, and Open Text Embedding \cite{neelakantan2022text}, which significantly advance the usage of text embeddings for general tasks. 

One major application of embedding models is neural retrieval \cite{lin2022pretrained}. By measuring the semantic relationship with the text embeddings, the relevant answers to the input query can be retrieved based on the embedding similarity. The most common form of embedding-based retrieval method is dense retrieval \cite{karpukhin-etal-2020-dense}, where the text encoder's outputs are aggregated (e.g., via [CLS] or mean-pooling) to compute the embedding similarity. Another common alternative is known as multi-vecor retrieval \cite{khattab2020colbert,Poly-encoders}, which applies fine-grained interactions for the text encoder's outputs to compute the embedding similarity. Finally, the text embeddings can also be transformed into term weights, which facilitates sparse or lexical retrieval \cite{luan-etal-2021-sparse,dai2020context,lin2021few}. Typically, the above retrieval methods are realized by different embedding models. To the best of our knowledge, no existing method is able to unify all these functionalities. 

Despite the substantial technical advancement, most of the existing text embeddings are developed only for English, where other languages are lagging behind. To mitigate this problem, continual efforts are presented from multiple directions. One is the development of pre-trained multi-lingual text encoders, such as mBERT \cite{pires-etal-2019-multilingual}, mT5 \cite{xue-etal-2021-mt5}, XLM-R \cite{conneau-etal-2020-unsupervised}. Another one is the curation of training and evaluation data for multi-lingual text embeddings, e.g., MIRACL \cite{zhang-etal-2023-miracl}, mMARCO \cite{bonifacio2021mmarco}, Mr. TyDi \cite{zhang-etal-2021-mr}, MKQA \cite{longpre-etal-2021-mkqa}. At the same time, the multi-lingual text embeddings are continually developed from the community, e.g., mDPR \cite{zhang2023toward}, mContriever \cite{mcontriever}, mE5 \cite{wang2022text}, etc. However, the current progress is still far from enough given the notable gap with English models and the huge imbalance between different languages.  

\section{M3-Embedding}
M3-Embedding realizes three-fold versatility. It supports a wide variety of languages and handles input data of different granularities. Besides, it unifies the common retrieval functionalities of text embeddings. Formally, given a query $q$ in an arbitrary language $x$, it is able to retrieve document $d$ in language $y$ from the corpus $D^y$: $d^y \leftarrow \mathrm{fn}^*(q^x, D^y)$. In this place, $\mathrm{fn}^*(\cdot)$ belongs to any of the functions: dense, lexical, or multi-vector retrieval; $y$ can be another language or the same language as $x$.

\begin{figure*}[t]
\centering
\includegraphics[width=1.0\textwidth]{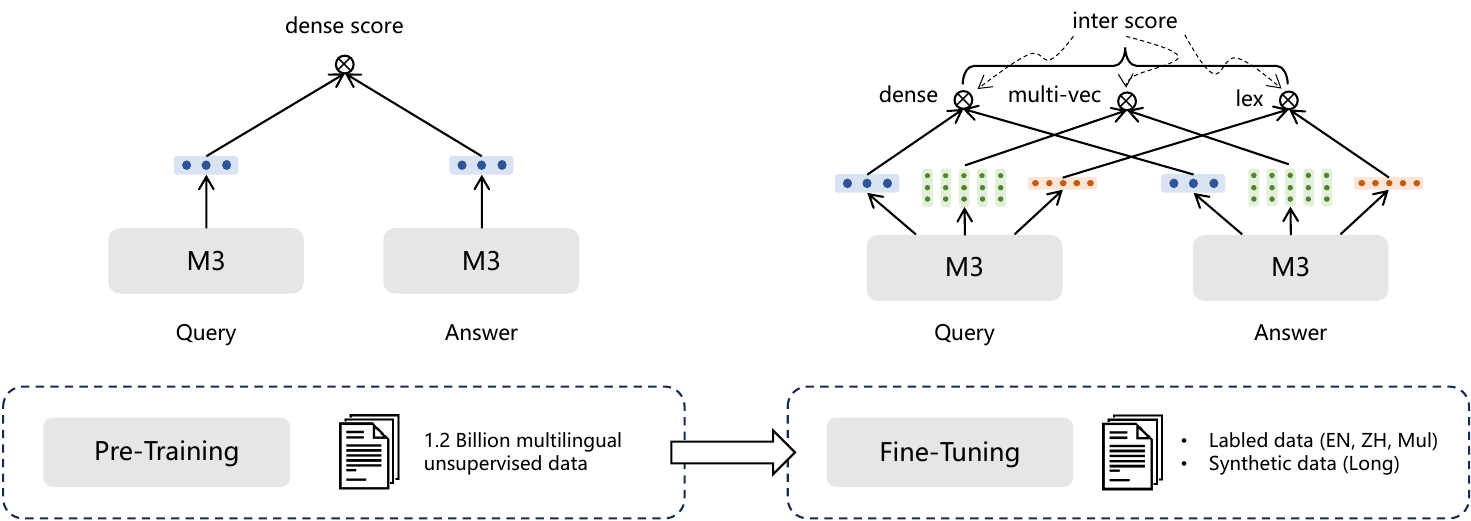}
\caption{\textbf{Multi-stage training process of M3-Embedding with self-knowledge distillation.}}
\label{fig:M3training}
\vspace{-10pt}
\end{figure*}

\subsection{Data Curation}
M3-Embedding calls for a large-scale and diverse multi-lingual dataset. 
In this work, we perform comprehensive data collection from three sources: the unsupervised data from unlabeled corpora, the fine-tuning data from labeled corpora, and the fine-tuning data via synthesization (shown as Table \ref{tab:data}). The three data sources complement to each other, which are applied to different stages of the training process. Particularly, the \textbf{unsupervised data} is curated by extracting the rich-semantic structures, e.g., title-body, title-abstract, instruction-output, etc., within a wide variety of multi-lingual corpora, including Wikipedia, S2ORC~\cite{lo-wang-2020-s2orc}, xP3~\cite{muennighoff-etal-2023-crosslingual}, mC4~\cite{2019t5}, CC-News~\cite{ccnews} and the well-curated data from MTP \cite{bge_embedding}. To learn the unified embedding space for cross-lingual semantic matching, the parallel sentences are introduced from two translation datasets, NLLB~\cite{nllb2022} and CCMatrix~\cite{schwenk-etal-2021-ccmatrix}. The raw data is filtered to remove potential bad contents and low-relevance samples. In total, it brings in \textit{1.2 billion} text pairs of \textit{194 languages} and \textit{2655 cross-lingual} correspondences. 

Besides, we collect relatively small but diverse and high-quality \textbf{fine-tuning data} from labeled corpora. For English, we incorporate 8 datasets, including HotpotQA \cite{yang-etal-2018-hotpotqa}, TriviaQA \cite{joshi-etal-2017-triviaqa}, NQ \cite{kwiatkowski-etal-2019-natural}, MS MARCO \cite{nguyen2016ms}, COLIEE \cite{kim2022coliee}, PubMedQA \cite{jin-etal-2019-pubmedqa}, SQuAD~\cite{rajpurkar-etal-2016-squad}, and NLI data from SimCSE \cite{gao-etal-2021-simcse}. For Chinese, we integrate 7 datasets, including DuReader \cite{he-etal-2018-dureader}, mMARCO-ZH \cite{bonifacio2021mmarco}, T$^2$-Ranking \cite{xie2023t2ranking}, LawGPT\cite{LAWGPT-zh}, CMedQAv2 \cite{cpubmedv2}, NLI-zh\footnote{\scriptsize \href{https://huggingface.co/datasets/shibing624/nli-zh-all}{https://huggingface.co/datasets/shibing624/nli-zh-all}}, and LeCaRDv2 \cite{li2023lecardv2}. For other languages, we leverage the training data from Mr. Tydi \cite{zhang-etal-2021-mr} and MIRACL \cite{zhang-etal-2023-miracl}. 

Finally, we generate \textbf{synthetic data} to mitigate the shortage of long document retrieval tasks and introduce extra multi-lingual fine-tuning data (denoted as MultiLongDoc). Specifically, we sample lengthy articles from Wikipedia, Wudao~\cite{yuan2021wudaocorpora} and mC4 datasets and randomly choose paragraphs from them. Then we use GPT-3.5 to generate questions based on these paragraphs. The generated question and the sampled article constitute a new text pair to the fine-tuning data. Detailed specifications are presented in Appendix~\ref{appendix:syn_data}.

\subsection{Hybrid Retrieval}
M3-Embedding unifies the common retrieval functionalities of the embedding model, i.e. dense retrieval, lexical (sparse) retrieval, and multi-vector retrieval. The formulation is presented as follows. 

$\bullet$ \textbf{Dense retrieval}. The input query $q$ is transformed into the hidden states $\mathbf{H_q}$ based on a text encoder. We use the normalized hidden state of the special token ``[CLS]'' for the representation of the query: $e_q = norm(\mathbf{H_q}[0])$. Similarly, we can get the embedding of passage $p$ as $e_p = norm(\mathbf{H_p}[0])$. Thus, the relevance score between query and passage is measured by the inner product between the two embeddings $e_q$ and $e_p$: $s_{dense} \leftarrow \langle e_p, e_q \rangle$. 

$\bullet$ \textbf{Lexical Retrieval}. The output embeddings are also used to estimate the importance of each term to facilitate lexical retrieval. For each term $t$ within the query (a term is corresponding to a token in our work), the term weight is computed as $w_{q_t} \leftarrow \mathsf{Relu}(\mathbf{W}_{lex}^T\mathbf{H_q}[i]))$, where $\mathbf{W}_{lex} \in \mathcal{R}^{d\times 1}$ is the matrix mapping the hidden state to a float number. If a term $t$ appears multiple times in the query, we only retain its max weight. We use the same way to compute the weight of each term in the passage. Based on the estimation term weights, the relevance score between query and passage is computed by the joint importance of the co-existed terms (denoted as $q \cap p$) within the query and passage: $s_{lex} \leftarrow \sum_{t\in q \cap p} (w_{q_t} * w_{p_t})$. 

$\bullet$ \textbf{Multi-Vector Retrieval}. As an extension of dense retrieval, the multi-vector method utilizes the entire output embeddings for the representation of query and passage: $E_q = norm(\mathbf{W}_{mul}^T\mathbf{H_q})$, $E_p = norm(\mathbf{W}_{mul}^T\mathbf{H_p})$, where $\mathbf{W}_{mul} \in \mathbb{R}^{d\times d}$ is the learnable projection matrix. Following ColBert~\cite{khattab2020colbert}, we use late-interaction to compute the fine-grained relevance score: $s_{mul} \leftarrow \frac{1}{N} \sum_{i=1}^{N} \max_{j=1}^{M} E_q[i] \cdot E_p^{T}[j]$; $N$ and $M$ are the lengths of query and passage. 

Thanks to the multi-functionality of the embedding model, the retrieval process can be conducted in a \textbf{hybrid process}. First of all, the candidate results can be individually retrieved by each of the methods (the multi-vector method can be exempted from this step due to its heavy cost). Then, the final retrieval result is re-ranked based on the integrated relevance score: 

\begin{equation}\label{eqn:hybrid_score}
    s_{rank} \leftarrow w_1 \cdot s_{dense} + w_2 \cdot s_{lex} + w_3 \cdot s_{mul}
\end{equation}
where the values of $w_1$, $w_2$ and $w_3$ depend on the downstream scenario.





\subsection{Self-Knowledge Distillation}
The embedding model is trained to discriminate the positive samples from the negative ones. For each of the retrieval methods, it is expected to assign a higher score for the query's positive samples compared with the negative ones. Therefore, the training process is conducted to minimize the InfoNCE loss, whose general form is presented by the following loss function:
\begin{equation}\label{eqn:loss}
     \mathcal{L}_{s(\cdot)} = - \log \frac{ \exp(s(q,p^*) / \tau) }{\sum_{p \in \{p^*, P'\}} \exp(s(q,p) / \tau) }.
\end{equation}
Here, $p^*$ and $P'$ stand for the positive and negative samples to the query $q$; $s(\cdot)$ is any of the functions within \{$s_{dense}(\cdot)$, $s_{lex}(\cdot)$, $s_{mul}(\cdot)$\}. 

The training objectives of different retrieval methods can be mutually conflicting with each their. Therefore, the native multi-objective training can be unfavorable to the embedding's quality. To facilitate the optimization of multiple retrieval functions, we propose to unify the training process on top of \textbf{self-knowledge distillation}. Particularly, based on the principle of ensemble learning \cite{buhlmann2012bagging}, the predictions from different retrieval methods can be integrated as a more accurate relevance score given their heterogeneous nature. In the simplest form, the integration can just be the 
weighted sum of different prediction scores: 
\begin{equation}
    s_{inter} \leftarrow w_1 \cdot s_{dense} + w_2 \cdot s_{lex} + w_3 \cdot s_{mul}. 
\end{equation}
Then we compute the weighted sum of $\mathcal{L}_{dense}$, $\mathcal{L}_{lex}$, $\mathcal{L}_{mul}$ and $\mathcal{L}_{inter}$ as the loss without self-knowledge distillation:
\begin{equation}
    \mathcal{L} \leftarrow \big(\lambda_1 \cdot \mathcal{L}_{dense} + \lambda_2 \cdot \mathcal{L}_{lex} + \lambda_3 \cdot \mathcal{L}_{mul} + \mathcal{L}_{inter}\big) / 4 . 
\end{equation}
In previous studies, the training quality of embedding model can benefit from knowledge distillation, which takes advantage of fine-grained soft labels from another ranking model \cite{hofstatter2021efficiently}. In this place, we simply employ the integration score $s_{inter}$ as the teacher, where the loss function of each retrieval method is modified as: 
\begin{equation}
    \mathcal{L}'_* \leftarrow - p(s_{inter}) * \log p(s_{*}). 
\end{equation}
Here, $p(\cdot)$ is the softmax activation; $s_{*}$ is any of the members within $s_{dense}$, $s_{lex}$, and $s_{mul}$. We further integrate and normalize the modified loss function: 
\begin{equation}
    \mathcal{L}' \leftarrow \big( \lambda_1 
 \cdot \mathcal{L}'_{dense} + \lambda_2 \cdot \mathcal{L}'_{lex} + \lambda_3 \cdot \mathcal{L}'_{mul}\big) / 3. 
\end{equation}
Finally, we derive the final loss function for self-knowledge distillation with the linear combination of $\mathcal{L}$ and $\mathcal{L}'$: $\mathcal{L}_{final} \leftarrow \big(\mathcal{L} + \mathcal{L}'\big) / 2$. 

The training process constitutes a \textbf{multi-stage workflow} (Figure \ref{fig:M3training}). 
In the first place, the text encoder (an XLM-RoBERTa~\cite{conneau-etal-2020-unsupervised} model adapted by RetroMAE~\cite{xiao-etal-2022-retromae} method) is pre-trained with the massive unsupervised data, where only the dense retrieval is trained in the basic form of contrastive learning. The self-knowledge distillation is applied to the second stage, where the embedding model is fine-tuned to establish the three retrieval functionalities. The random initialization of $\mathbf{W}_{lex}$ led to poor $s_{lex}$ accuracy and high $\mathcal{L}_{lex}$ at the beginning of the training. In order to reduce the impact of this, we set $w_1=1$, $w_2=0.3$, $w_3=1$, $\lambda_1=1$, $\lambda_2=0.1$ and $\lambda_3=1$ during the training process. Both labeled and synthetic data are used in this stage, where hard negative samples are introduced for each query following the ANCE method \cite{xiong2020approximate}. (See Appendix~\ref{appendix_sec:hyperparameters} for more details.)

\begin{figure}[t]
\centering
\includegraphics[width=1.0\linewidth]{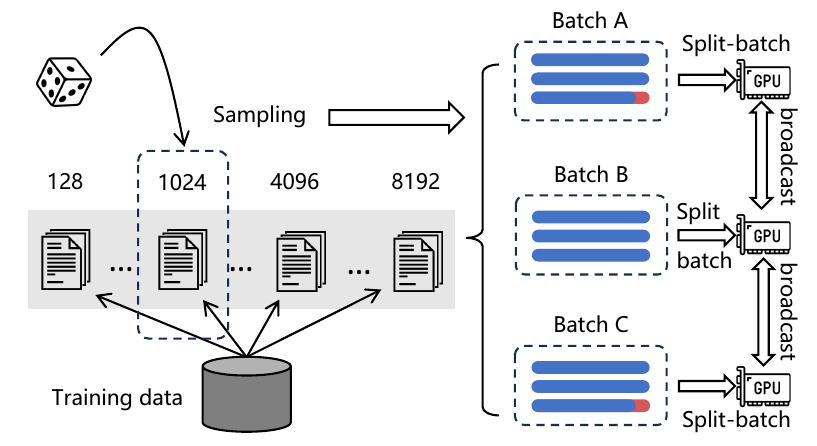}
\caption{\textbf{Efficient Batching.} (Data is grouped and sampled by length. Gradient-checkpointing and cross-GPU broadcasting are enabled to save memory.)}
\label{fig:batch}
\vspace{-10pt}
\end{figure}

\renewcommand{\arraystretch}{1.1}
\begin{table*}[!t]
    \centering
    \footnotesize
    \setlength{\tabcolsep}{2pt}
    \begin{tabular}{l|c|cccccccccccccccccc}
    \hline
    Model & Avg & ar & bn & en & es & fa & fi & fr & hi & id & ja & ko & ru & sw & te & th & zh & de & yo \\
    \hline
    \multicolumn{20}{l}{Baselines~(\textit{Prior Work})} \\
    \hline
    BM25 & 31.9 & 39.5 & 48.2 & 26.7 & 7.7 & 28.7 & 45.8 & 11.5 & 35.0 & 29.7 & 31.2 & 37.1 & 25.6 & 35.1 & 38.3 & 49.1 & 17.5 & 12.0 & 56.1 \\
    mDPR & 41.8 & 49.9 & 44.3 & 39.4 & 47.8 & 48.0 & 47.2 & 43.5 & 38.3 & 27.2 & 43.9 & 41.9 & 40.7 & 29.9 & 35.6 & 35.8 & 51.2 & 49.0 & 39.6  \\
    mContriever & 43.1 & 52.5 & 50.1 & 36.4 & 41.8 & 21.5 & 60.2 & 31.4 & 28.6 & 39.2 & 42.4 & 48.3 & 39.1 & 56.0 & 52.8 & 51.7 & 41.0 & 40.8 & 41.5 \\
    mE5$_{\mathrm{{\text{large}}}}$ & 66.6 & 76.0 & 75.9 & 52.9 & 52.9 & 59.0 & 77.8 & 54.5 & 62.0 & 52.9 & 70.6 & 66.5 & 67.4 & 74.9 & 84.6 & 80.2 & 56.0 & 56.4 & 78.3 \\
    E5$_{\mathrm{\text{mistral-7b}}}$ & 63.4 & 73.3 & 70.3 & 57.3 & 52.2 & 52.1 & 74.7 & 55.2 & 52.1 & 52.7 & 66.8 & 61.8 & 67.7 & 68.4 & 73.9 & 74.0 & 54.0 & 54.1 & 79.7 \\
    OpenAI-3 & 54.9 & - & - & - & - & - & - & - & - & - & - & - & - & - & - & - & - & - & - \\
    \hline
    \multicolumn{20}{l}{M3-Embedding~(\textit{Our Work})} \\
    \hline
    Dense & 69.2 & 78.4 & 80.0 & 56.9 & 56.1 & 60.9 & 78.6 & 58.3 & 59.5 & 56.1 & 72.8 & 69.9 & 70.1 & 78.7 & 86.2 & 82.6 & 62.7 & 56.7 & 81.8 \\
    Sparse & 53.9 & 67.1 & 68.9 & 43.8 & 38.6 & 45.1 & 65.4 & 35.3 & 48.2 & 48.9 & 56.1 & 61.5 & 44.5 & 57.9 & 79.1 & 70.9 & 36.1 & 32.5 & 70.0 \\
    Multi-vec & 70.5 & 79.6 & 81.0 & 59.3 & 57.8 & 62.0 & 80.1 & 59.4 & 61.5 & 58.3 & 74.5 & 71.2 & 71.2 & 79.1 & 87.9 & 83.0 & 63.7 & 58.0 & 82.4 \\
    Dense+Sparse & 70.4 & 79.6 & 80.7 & 58.8 & 58.1 & 62.3 & 79.7 & 58.0 & 62.9 & 58.3 & 73.9 & 71.2 & 69.8 & 78.5 & 87.2 & 83.1 & 63.5 & 57.7 & 83.3 \\
    All & \textbf{71.5} & \textbf{80.2} & \textbf{81.5} & \textbf{59.6} & \textbf{59.7} & \textbf{63.4} & \textbf{80.4} & \textbf{61.2} & \textbf{63.3} & \textbf{59.0} & \textbf{75.2} & \textbf{72.1} & \textbf{71.7} & \textbf{79.6} & \textbf{88.1} & \textbf{83.7} & \textbf{64.9} & \textbf{59.8} & \textbf{83.5} \\
    \hline
    
    \end{tabular}
    \vspace{-5pt}
    \caption{\textbf{Multi-lingual retrieval performance on the MIRACL dev set} (measured by nDCG@10).}
    \vspace{-10pt}
    \label{tab:miracl_ndcg_results}
\end{table*}

\subsection{Efficient Batching} 
The embedding model needs to learn from diverse and massive multi-lingual data to fully capture the general semantic of different languages. It also needs to keep the batch size as large as possible (introducing a huge amount of in-batch negatives) to ensure the discriminativeness of text embeddings. Given the limitations on GPU's memory and computation power, people usually truncate the input data into short sequences for high throughput of training and a large batch size. 
However, the common practice is not a feasible option for M3-Embedding because it needs to learn from both short and long-sequence data to effectively handle the input of different granularities. In our work, we improve the training efficiency by optimizing the batching strategy, which enables high training throughput and large batch sizes. 

Particularly, the training data is pre-processed by being grouped by sequence length. When producing a mini-batch, the training instances are sampled from the same group. Due to the similar sequence lengths, it significantly reduces sequence padding (Figure~\ref{fig:batch}, marked in red) and facilitates a more effective utilization of GPUs. 
Besides, when sampling the training data for different GPUs, the random seed is always fixed, which ensures the load balance and minimizes the waiting time in each training step. 
Besides, when handling long-sequence training data, the mini-batch is further divided into sub-batches, which takes less memory footprint. We iteratively encode each sub-batch using gradient checkpointing \cite{chen2016training} and gather all generated embeddings. 
This method can significantly increase the batch size.
For example, when processing text with a length of 8192, the batch size can be increased by more than 20 times. (see Appendx~\ref{appendix_sec:efficient_batching} for more details.)
Finally, the embeddings from different GPUs are broadcasted, allowing each device to obtain all embeddings in the distributed environment, which notably expands the scale of in-bath negative samples. 


For users who are severely limited in computation or data resource, we present an even simpler method called MCLS (Multi-CLS), which simply inserts multiple CLS tokens to the long document during inference, and takes the average of all CLS embeddings as the ultimate embedding of the document. Despite simplicity, it is surprisingly effective in practice. 
 (See Appendix~\ref{appendix:mcls} for more details.)

\section{Experiment}

In this section, we investigate M3-Embedding's performance in terms of multi-lingual retrieval, cross-lingual retrieval, and long-doc retrieval. We also explore the impact of its technical factors. 


\subsection{Multi-Lingual Retrieval}
We evaluate the multi-lingual retrieval performance with MIRACL~\cite{zhang-etal-2023-miracl}, which consists of ad-hoc retrieval tasks in 18 languages. Each task is made up of query and passage presented in the same language. Following the official benchmark, we evaluate our method using Pyserini~\cite{Pyserini}, and use nDCG@10 as the primary evaluation metric (Recall@100 is also measured and reported in Appendix~\ref{appendix_sec:additional_results}). Specifically, for the dense method (denoted as \textit{\underline{Dense}}), we first use it to generate the embeddings of the corpus and then build the dense index for searching top-1000 candidates with Faiss. For the sparse method (denoted as \textit{\underline{Sparse}}), we first use it to generate the weights of the corpus and then build the sparse index for searching top-1000 candidates with Lucene. For the multi-vector method (denoted as \textit{\underline{Multi-vec}}), considering its heavy cost, we use it as reranker to re-rank the top-200 candidates from dense method. For the hybrid retrieval of dense method and sparse method (denoted as \textit{\underline{Dense+Sparse}}), we set $w_1=1$, $w_2=0.3$ and $w_3=0$ in equation\eqref{eqn:hybrid_score} to re-rank the union set of top-1000 candidates from Dense and top-1000 candidate from Sparse. For the hybrid retrieval of all three methods (denoted as \textit{\underline{All}}), we set $w_1=1$, $w_2=0.3$ and $w_3=1$ in equation\eqref{eqn:hybrid_score} to re-rank the top-200 candidates from Dense. 

We incorporate the following baselines in our experiment: the lexical retrieval method: BM25~\cite{BM25_1}; the dense retrieval methods: mDPR\footnote{\scriptsize \href{https://huggingface.co/castorini/mdpr-tied-pft-msmarco}{https://huggingface.co/castorini/mdpr-tied-pft-msmarco}}~\cite{zhang2023toward}, mContriever\footnote{\scriptsize \href{https://huggingface.co/facebook/mcontriever-msmarco}{https://huggingface.co/facebook/mcontriever-msmarco}}~\cite{mcontriever}, mE5$_{\mathrm{\text{large}}}$~\cite{wang2022text} and E5$_{\mathrm{\text{mistral-7b}}}$~\cite{e5-mistral-7b-instruct}. 
To make the BM25 and M3 more comparable, in the experiment, we use the same tokenizer as M3 (i.e., the tokenizer of XLM-Roberta) for BM25. 
Using the same vocabulary from XLM-Roberta can also ensure that both approaches have the same retrieval latency. The results of BM25 with different tokenizers are shown in Appendix~\ref{appendix:tokenizer_bm25}. 
We also make a comparison with Text-Embedding-3-Large(abbreviated as OpenAI-3), which was recently released by OpenAI\footnote{\scriptsize \href{https://platform.openai.com/docs/guides/embeddings}{https://platform.openai.com/docs/guides/embeddings}}. 

\renewcommand{\arraystretch}{1.0}
\begin{table*}[t]
    \centering
    \footnotesize
    \setlength{\tabcolsep}{3pt}
    \begin{tabular}{lcccccc|ccccc}
    \hline
    & \multicolumn{6}{c|}{Baselines~(\textit{Prior Work})} & \multicolumn{5}{c}{M3-Embedding~(\textit{Our Work})} \\
    \hline
    & BM25 & mDPR & mContriever & mE5$_{\mathrm{{\text{large}}}}$ & E5$_{\mathrm{\text{mistral-7b}}}$ & OpenAI-3 & Dense & Sparse & Multi-vec & Dense+Sparse & All \\
    \hline
    ar      & 18.9 & 48.2 & 58.2 & 68.7 & 59.6 & 65.6 & 71.1 & 23.5 & 71.4 & 71.1 & \textbf{71.5} \\
    da      & 49.3 & 67.4 & 73.9 & 77.4 & \textbf{77.8} & 73.6 & 77.2 & 55.4 & 77.5 & 77.4 & 77.6 \\
    de      & 35.4 & 65.8 & 71.7 & 76.9 & \textbf{77.0} & 73.6 & 76.2 & 43.3 & 76.3 & 76.4 & 76.3 \\
    es      & 43.4 & 66.8 & 72.6 & 76.4 & \textbf{77.4} & 73.9 & 76.4 & 50.6 & 76.6 & 76.7 & 76.9 \\
    fi      & 46.3 & 56.2 & 70.2 & 74.0 & 72.0 & 72.7 & 75.1 & 51.1 & 75.3 & 75.3 & \textbf{75.5} \\
    fr      & 45.3 & 68.2 & 72.8 & 75.5 & \textbf{78.0} & 74.1 & 76.2 & 53.9 & 76.4 & 76.6 & 76.6 \\
    he      & 26.9 & 49.7 & 63.8 & 69.6 & 47.2 & 58.1 & 72.4 & 31.1 & 72.9 & 72.5 & \textbf{73.0} \\
    hu      & 38.2 & 60.4 & 69.7 & 74.7 & \textbf{75.0} & 71.2 & 74.7 & 44.6 & 74.6 & 74.9 & \textbf{75.0} \\
    it      & 45.2 & 66.0 & 72.3 & 76.8 & \textbf{77.1} & 73.6 & 76.0 & 52.5 & 76.4 & 76.3 & 76.5 \\
    ja      & 24.5 & 60.3 & 64.8 & 71.5 & 65.1 & 71.9 & 75.0 & 31.3 & 75.1 & 75.0 & \textbf{75.2} \\
    km      & 27.8 & 29.5 & 26.8 & 28.1 & 34.3 & 33.9 & 68.6 & 30.1 & 69.1 & 68.8 & \textbf{69.2} \\
    ko      & 27.9 & 50.9 & 59.7 & 68.1 & 59.4 & 63.9 & 71.6 & 31.4 & 71.7 & 71.6 & \textbf{71.8} \\
    ms      & 55.9 & 65.5 & 74.1 & 76.3 & 77.2 & 73.3 & 77.2 & 62.4 & \textbf{77.4} & \textbf{77.4} & \textbf{77.4} \\
    nl      & 56.2 & 68.2 & 73.7 & 77.8 & \textbf{79.1} & 74.2 & 77.4 & 62.4 & 77.6 & 77.7 & 77.6 \\
    no      & 52.1 & 66.7 & 73.5 & 77.3 & 76.6 & 73.3 & 77.1 & 57.9 & 77.2 & \textbf{77.4} & 77.3 \\
    pl      & 40.8 & 63.3 & 71.6 & 76.7 & \textbf{77.1} & 72.7 & 76.3 & 46.1 & 76.5 & 76.3 & 76.6 \\
    pt      & 44.9 & 65.5 & 72.0 & 73.5 & \textbf{77.5} & 73.7 & 76.3 & 50.9 & 76.4 & 76.5 & 76.4 \\
    ru      & 33.2 & 62.7 & 69.8 & \textbf{76.8} & 75.5 & 72.0 & 76.2 & 36.9 & 76.4 & 76.2 & 76.5 \\
    sv      & 54.6 & 66.9 & 73.2 & 77.6 & \textbf{78.3} & 74.0 & 76.9 & 59.6 & 77.2 & 77.4 & 77.4 \\
    th      & 37.8 & 53.8 & 66.9 & 76.0 & 67.4 & 65.2 & 76.4 & 42.0 & 76.5 & 76.5 & \textbf{76.6} \\
    tr      & 45.8 & 59.1 & 71.1 & 74.3 & 73.0 & 71.8 & 75.6 & 51.8 & 75.9 & \textbf{76.0} & \textbf{76.0} \\
    vi      & 46.6 & 63.4 & 70.9 & 75.4 & 70.9 & 71.1 & 76.6 & 51.8 & 76.7 & 76.8 & \textbf{76.9} \\
    zh\_cn  & 31.0 & 63.7 & 68.1 & 56.6 & 69.3 & 70.7 & 74.6 & 35.4 & 74.9 & 74.7 & \textbf{75.0} \\
    zh\_hk  & 35.0 & 62.8 & 68.0 & 58.1 & 65.1 & 69.6 & 73.8 & 39.8 & 74.1 & 74.0 & \textbf{74.3} \\
    zh\_tw  & 33.5 & 64.0 & 67.9 & 58.1 & 65.8 & 69.7 & 73.5 & 37.7 & 73.5 & \textbf{73.6} & \textbf{73.6} \\
    \hline
    Avg     & 39.9 & 60.6 & 67.9 & 70.9 & 70.1 & 69.5 & 75.1 & 45.3 & 75.3 & 75.3 & \textbf{75.5} \\
    \hline
    \end{tabular}
    \vspace{-5pt}
    \caption{\textbf{Cross-lingual retrieval performance on MKQA} (measured by Recall@100).}
    \vspace{-5pt}
    \label{tab:mkqa_recall@100_results}
\end{table*}

We can make the following observations according to the experiment result in Table \ref{tab:miracl_ndcg_results}. Firstly, M3-Embedding already achieves a superior retrieval performance with only its dense retrieval functionality (Dense). It not only outperforms other baseline methods in the average performance, but also maintains a consistent empirical advantage in most of individual languages. Even compared with E5$_{\mathrm{\text{mistral-7b}}}$, which leverages a much larger Mistral-7B model as the text encoder and specifically trained with English data, our method is able to produce a similar result in English and notably higher results in the other languages. Besides, the sparse retrieval functionality (Sparse) is also effectively trained by M3-Embedding, as it outperforms the typical BM25 methods in all languages. We can also observe the additional improvement from multi-vector retrieval, which relies on fine-grained interactions between query and passage's embeddings to compute the relevance score. Finally, the collaboration of dense and sparse method (Dense+Sparse) leads to a further improvement over each individual method, and the collaboration of all three methods (All) brings forth the best performance.

\begin{table*}[!t]
    \centering
    \small
    \setlength{\tabcolsep}{2.5pt}
    \begin{tabular}{lc|c|ccccccccccccc}
    \hline
    & Max Length & Avg & ar & de & en & es & fr & hi & it & ja & ko & pt & ru & th & zh \\
    \hline
    \multicolumn{16}{l}{Baselines~(\textit{Prior Work})} \\
    \hline
    BM25 & 8192 & 53.6 & 45.1 & 52.6 & 57.0 & 78.0 & 75.7 & 43.7 & 70.9 & 36.2 & 25.7 & 82.6 & 61.3 & 33.6 & 34.6 \\
    mDPR & 512 & 23.5 & 15.6 & 17.1 & 23.9 & 34.1 & 39.6 & 14.6 & 35.4 & 23.7 & 16.5 & 43.3 & 28.8 & 3.4 & 9.5 \\
    mContriever & 512 & 31.0 & 25.4 & 24.2 & 28.7 & 44.6 & 50.3 & 17.2 & 43.2 & 27.3 & 23.6 & 56.6 & 37.7 & 9.0 & 15.3 \\
    mE5$_{\mathrm{{\text{large}}}}$ & 512 & 34.2 & 33.0 & 26.9 & 33.0 & 51.1 & 49.5 & 21.0 & 43.1 & 29.9 & 27.1 & 58.7 & 42.4 & 15.9 & 13.2 \\
    E5$_{\mathrm{\text{mistral-7b}}}$ & 8192 & 42.6 & 29.6 & 40.6 & 43.3 & 70.2 & 60.5 & 23.2 & 55.3 & 41.6 & 32.7 & 69.5 & 52.4 & 18.2 & 16.8 \\
    text-embedding-ada-002 & 8191 & 32.5 & 16.3 & 34.4 & 38.7 & 59.8 & 53.9 & 8.0 & 46.5 & 28.6 & 20.7 & 60.6 & 34.8 & 9.0 & 11.2 \\
    jina-embeddings-v2-base-en & 8192 & - & - & - & 37.0 & - & - & - & - & - & - & - & - & - & - \\
    \hline
    \multicolumn{16}{l}{M3-Embedding~(\textit{Our Work})} \\
    \hline
    Dense           & 8192 & 52.5 & 47.6 & 46.1 & 48.9 & 74.8 & 73.8 & 40.7 & 62.7 & 50.9 & 42.9 & 74.4 & 59.5 & 33.6 & 26.0 \\
    Sparse          & 8192 & 62.2 & 58.7 & 53.0 & 62.1 & 87.4 & 82.7 & 49.6 & 74.7 & 53.9 & 47.9 & 85.2 & 72.9 & 40.3 & 40.5 \\
    Multi-vec    & 8192 & 57.6 & 56.6 & 50.4 & 55.8 & 79.5 & 77.2 & 46.6 & 66.8 & 52.8 & 48.8 & 77.5 & 64.2 & 39.4 & 32.7 \\
    Dense+Sparse    & 8192 & 64.8 & 63.0 & 56.4 & \textbf{64.2} & \textbf{88.7} & \textbf{84.2} & \textbf{52.3} & \textbf{75.8} & 58.5 & 53.1 & \textbf{86.0} & \textbf{75.6} & 42.9 & \textbf{42.0} \\
    All             & 8192 & \textbf{65.0} & \textbf{64.7} & \textbf{57.9} & 63.8 & 86.8 & 83.9 & 52.2 & 75.5 & \textbf{60.1} & \textbf{55.7} & 85.4 & 73.8 & \textbf{44.7} & 40.0 \\
    \hline
    \multicolumn{16}{l}{M3-w.o.long} \\
    \hline
    Dense-w.o.long & 8192 & 41.2 & 35.4 & 35.2 & 37.5 & 64.0 & 59.3 & 28.8 & 53.1 & 41.7 & 29.8 & 63.5 & 51.1 & 19.5 & 16.5 \\
    Dense-w.o.long (MCLS) & 8192 & 45.0 & 37.9 & 43.3 & 41.2 & 67.7 & 64.6 & 32.0 & 55.8 & 43.4 & 33.1 & 67.8 & 52.8 & 27.2 & 18.2 \\
    \hline
    \end{tabular}
    \vspace{-5pt}
    \caption{\textbf{Evaluation of multilingual long-doc retrieval on the MLDR test set} (measured by nDCG@10).}
    \vspace{-15pt}
    \label{tab:bge_long_results}
\end{table*}

\subsection{Cross-Lingual Retrieval}
We make evaluation for the cross-lingual retrieval performance with the MKQA benchmark~\cite{longpre-etal-2021-mkqa}, which includes queries in 25 non-English languages. For each query, it needs to retrieve the passages containing answers from the English Wikipedia corpus. In our experiment, we make use of the well-processed corpus offered by the BEIR\footnote{\scriptsize \href{https://huggingface.co/datasets/BeIR/nq}{https://huggingface.co/datasets/BeIR/nq}}~\cite{thakur2021beir}. Following the previous study~\cite{mcontriever}, we report Recall@100 as the primary metric (Recall@20 is reported as an auxiliary metric in the Appendix~\ref{appendix_sec:additional_results}). For Dense+Sparse method and All method, we set the same weights as in MIRACL dataset.

The experiment result is shown in Table \ref{tab:mkqa_recall@100_results}. Similar to our observation in multi-lingual retrieval, M3-Embedding continues to produce a superior performance, where it notably outperforms other baseline methods purely with its dense retrieval functionality (Dense). The collaboration of different retrieval methods brings in further improvements, leading to the best empirical performance of cross-lingual retrieval. Besides, we can also observe the following interesting results which are unique to this benchmark. Firstly, the performance gaps are not as significant as MIRACL, where competitive baselines like E5$_{\mathrm{\text{mistral-7b}}}$ is able to produce similar or even better results on some of the testing languages. However, the baselines are prone to bad performances in many other languages, especially the low-resource languages, such as ar, km, he, etc. In contrast, M3-Embedding maintains relatively stable performances in all languages, which can largely be attributed to its pre-training over comprehensive unsupervised data. Secondly, although M3-Embedding (Sparse) is still better than BM25, it performs badly compared with other methods. This is because there are only very limited co-existed terms for cross-lingual retrieval as the query and passage are presented in different languages.

\subsection{Multilingual Long-Doc Retrieval}
We evaluate the retrieval performance with longer sequences with two benchmarks: MLDR (Multilingual Long-Doc Retrieval), which is curated by the multilingual articles from Wikipedia, Wudao and mC4 (see Table~\ref{tab:bge_long}), and NarrativeQA~\cite{kocisky-etal-2018-narrativeqa, jina}, which is only for English. In addition to the previous baselines, we further introduce JinaEmbeddingv2~\cite{jina}, text-embedding-ada-002 and text-embedding-3-large from OpenAI given their outstanding long-doc retrieval capability. For Dense+Sparse method, we set $w_1=0.2$, $w_2=0.8$ and $w_3=0$ in equation\eqref{eqn:hybrid_score}. For All method, we set $w_1=0.15$, $w_2=0.5$ and $w_3=0.35$ in equation\eqref{eqn:hybrid_score}.

The evaluation result on MLDR is presented in Table~\ref{tab:bge_long_results}. 
Interestingly, M3 (Sparse) turns out to be a more effective method for long document retrieval, which achieves another about 10 points improvement over the dense method. Besides, the multi-vector retrieval is also impressive, which brings 5.1+ points improvement over M3 (Dense). Finally, the combination of different retrieval methods leads to a remarkable average performance of 65.0. 


To explore the reason for M3-Embedding's competitiveness in long-document retrieval, we perform the ablation study by removing the long document data from the fine-tuning stage (denoted as w.o. long). After this modification, the dense method, i.e. Dense-w.o.long, can still outperform the majority of baselines, which indicates that its empirical advantage has been well established during the pre-training stage. We also propose a simple strategy, MCLS, to address this situation (no data or no GPU resource for document-retrieval fine-tuning). Experimental results indicate that MCLS can significantly improve the performance of document retrieval without training ($41.2 \to 45.0$). 

\begin{table}[]
    \centering
    \small
    \setlength{\tabcolsep}{4pt}
    \begin{tabular}{lcc}
    \hline
    Model & Max Length & nDCG@10 \\
    \hline
    \multicolumn{3}{l}{Baselines~(\textit{Prior Work})} \\
    \hline
    mDPR                                    & 512 & 16.3 \\
    mContriever                             & 512 & 23.3 \\
    mE5$_{\mathrm{{\text{large}}}}$         & 512 & 24.2 \\
    E5$_{\mathrm{\text{mistral-7b}}}$       & 8192 & 49.9 \\
    text-embedding-ada-002                  & 8191 & 41.1 \\
    text-embedding-3-large                  & 8191 & 51.6 \\
    jina-embeddings-v2-base-en              & 8192 & 39.4 \\
    \hline
    \multicolumn{3}{l}{M3-Embedding~(\textit{Our Work})} \\
    \hline
    Dense                                   & 8192 & 48.7 \\
    Sparse                                  & 8192 & 57.5 \\
    Multi-vec                            & 8192 & 55.4 \\
    Dense+Sparse                            & 8192 & 60.1 \\
    All                                     & 8192 & \textbf{61.7} \\
    \hline
    \end{tabular}
    \caption{\textbf{Evaluation on NarrativeQA} (nDCG@10).}
    \label{tab:narrativeqa_results}
\end{table}

We make further analysis with NarrativeQA (Table~\ref{tab:narrativeqa_results}), where we can make a similar observation as MLDR. Besides, with the growth of sequence length, our method gradually expands its advantage over baseline methods (Figure \ref{fig:narrativeqa_trend}), which reflects its proficiency in handling long inputs.

\subsection{Ablation study} \label{ablation_study} 

\noindent \textbf{Self-knowledge distillation}. The ablation study is performed to analyze the impact of self-knowledge distillation (skd). Particularly, we disable the distillation processing and have each retrieval method trained independently (denoted as M3-w.o.skd). According to our evaluation on MIRACL (Table~\ref{tab:brief_unify_ablation}), the original method, i.e. M3-w.skd, is able to achieve better performances than the ablation method in all settings, i.e., Dense, Sparse, Multi-vec. Notably, the impact is more pronounced for sparse retrieval. Such a result also reflects the incompatibility between dense and sparse retrieval methods. With skd, the incompatibility can be largely overcome. (More detailed results are available in Appendix~\ref{appendix_sec:additional_results}.)

\begin{table}[]
    \centering
    \small
    \setlength{\tabcolsep}{3pt}
    \begin{tabular}{ll|c}
        \hline
        \multicolumn{2}{l|}{Model} & MIRACL  \\
        \hline
        \multirow{3}{*}{M3-w.skd} & Dense & 69.2 \\
        & Sparse & 53.9 \\
        & Multi-vec & 70.5 \\
        \hline
        \multirow{3}{*}{M3-w.o.skd} & Dense & 68.7 \\
        & Sparse & 36.7 \\
        & Multi-vec & 69.3 \\
        \hline
    \end{tabular}
    \caption{\textbf{Ablation study of self-knowledge distillation on the MIRACL dev set} (nDCG@10).}
    \label{tab:brief_unify_ablation}
\end{table}

\begin{table}[]
    \centering
    \small
    \setlength{\tabcolsep}{3pt}
    \begin{tabular}{l|cc}
        \hline
        Model (Dense) & MIRACL  \\
        \hline
        Fine-tune & 60.5  \\
        RetroMAE + Fine-tune & 66.1  \\
        RetroMAE + Unsup + Fine-tune & 69.2 \\
        \hline
    \end{tabular}
    \caption{\textbf{Ablation study of multi-stage training on the MIRACL dev set} (nDCG@10).}
    \label{tab:stage}
    \vspace{-8pt}
\end{table}

\noindent \textbf{Impact of multi-stage training}. We also make explorations for the impacts from different training stages. \textit{Fine-tuning} indicates the direct fine-tuning from XLM-RoBERTA~\cite{xlm-roberta}; \textit{RetroMAE+Fine-tuning} refers to the fine-tuning on the pre-trained model from RetroMAE~\cite{xiao-etal-2022-retromae}. Meanwhile, \textit{RetroMAE+Unsup+Fine-tuning} involves fine-tuning on a model that is trained with RetroMAE and then pre-trained on unsupervised data.
The results are presented in Table~\ref{tab:stage}.
We can observe that RetroMAE can significantly improve the retrieval performance, and pre-training on unsupervised data can further enhance the retrieval quality of the embedding model. (More detailed results are available in Appendix~\ref{appendix_sec:additional_results}.)

\section{Conclusion}
In this paper, we introduce M3-Embedding, which substantially advances the versatility of text embeddings in terms of supporting multi-lingual retrieval, handling input of diverse granularities, and unifying different retrieval functionalities. M3-Embedding presents three technical contributions: self-knowledge distillation, efficient batching, and high-quality curation of data. The effectiveness of M3-Embedding is empirically verified, where it leads to superior performances on multi-lingual retrieval, cross-lingual retrieval, and multi-lingual long-document retrieval tasks. 


\section*{Limitations}

First of all, while our proposed M3-Embedding model achieves state-of-the-art performance on popular multi-lingual and cross-lingual benchmarks such as MIRACL and MKQA, it is important to acknowledge that the generalizability of our approach to diverse datasets and real-world scenarios needs to be further investigated. Different datasets may have varying characteristics and challenges that could affect the performance of our model. Secondly, while M3-Embedding is designed to process inputs of different granularities, including long documents of up to 8192 tokens, we acknowledge that processing extremely long documents could pose challenges in terms of computational resources and model efficiency. The performance of our model on very long documents or documents exceeding the specified token limit needs to be further investigated. Furthermore, we claim support for more than 100 working languages in M3-Embedding. However, the potential variations in performance across different languages are not thoroughly discussed. Further analysis and evaluation on a broader range of languages are necessary to understand the robustness and effectiveness of our model across different language families and linguistic characteristics.

\section*{Ethics Consideration}

Our work proposes a new embedding model called M3-Embedding, which is distingulished for its versality in multi-linguality, multi-functionality and multi-granularity. Because our model will be publicly avaliable, it is influenced by the inherent impacts of open-source model. Moreover, we use the multilingual data including all kinds of languages in the training of M3-Embedding. However, due to the uneven distribution of training data for different languages, the model's performance may vary across languages, which could potentially be seen as discriminatory or unfair. We ensure that our work is conformant to the ACL Ethics Policy\footnote{\scriptsize \href{https://www.aclweb.org/portal/content/acl-code-ethics}{https://www.aclweb.org/portal/content/acl-code-ethics}}.

\section*{Acknowledgements}

We would like to thank anonymous reviewers for their helpful feedback, and ACL 2024 and ACL Rolling Review organizers for their efforts. This research is supported by National Science and Technology Major Project (2023ZD0121504).

\bibliography{anthology, custom}

\clearpage
\appendix

\begin{table*}[!t]
    \centering
    \footnotesize
    \begin{tabular}{ccccccc}
    \hline
    Language & Source  & \#train & \#dev & \#test & \#cropus & Avg. Length of Docs \\
    \hline
    ar & Wikipedia & 1,817 & 200 & 200 & 7,607 & 9,428 \\
    de & Wikipedia, mC4 & 1,847 & 200 & 200 & 10,000 & 9,039 \\
    en & Wikipedia & 10,000 & 200 & 800 & 200,000 & 3,308 \\
    es & Wikipedia, mC4 & 2,254 & 200 & 200 & 9,551 & 8,771 \\
    fr & Wikipedia & 1,608 & 200 & 200 & 10,000 & 9,659 \\
    hi & Wikipedia & 1,618 & 200 & 200 & 3,806 & 5,555 \\
    it & Wikipedia & 2,151 & 200 & 200 & 10,000 & 9,195 \\
    ja & Wikipedia & 2,262 & 200 & 200 & 10,000 & 9,297 \\
    ko & Wikipedia & 2,198 & 200 & 200 & 6,176 & 7,832 \\
    pt & Wikipedia & 1,845 & 200 & 200 & 6,569 & 7,922 \\
    ru & Wikipedia & 1,864 & 200 & 200 & 10,000 & 9,723 \\
    th & mC4 & 1,970 & 200 & 200 & 10,000 & 8,089 \\
    zh & Wikipedia, Wudao & 10,000 & 200 & 800 & 200,000 & 4,249 \\
    \hline
    Total & -- & 41,434 & 2,600 & 3,800 & 493,709 & 4,737 \\
    \hline
    \end{tabular}
    \caption{Specifications of MultiLongDoc dataset.}
    \label{tab:bge_long}
\end{table*} 

\section{Details of Datasets}
\subsection{Collected Data}

\renewcommand{\arraystretch}{1.4}
\begin{table}[h]
    \centering
    \footnotesize
    \begin{tabular}{C{2.4cm}|C{2.2cm}|C{1.2cm}}
    \toprule
        Data Source & Language & Size \\
        \hline
        \multicolumn{3}{c}{Unsupervised Data} \\
        \hline
        MTP & EN, ZH & 291.1M \\
        \hline
        S2ORC, Wikipeida & EN & 48.3M \\
        \hline
        xP3, mC4, CC-News & Multi-Lingual & 488.4M \\
        \hline
        NLLB, CCMatrix & Cross-Lingual & 391.3M \\
        \hline
        CodeSearchNet & Text-Code & 344.1K \\
        \hline
        Total & -- & 1.2B \\ 
        \hline
        \hline
        \multicolumn{3}{c}{Fine-tuning Data} \\
        \hline
        MS MARCO, HotpotQA, NQ, NLI, etc. & EN & 1.1M \\
        \hline
        DuReader, T$^2$-Ranking, NLI-zh, etc. & ZH & 386.6K \\
        \hline
        MIRACL, Mr.TyDi & Multi-Lingual & 88.9K \\
        \hline
        \hline
        MultiLongDoc & Multi-Lingual & 41.4K \\
    \bottomrule
    \end{tabular}
    \caption{Specification of training data.}
    \vspace{-10pt}
    \label{tab:data}
\end{table}

\begin{figure*}[!t]
\centering
\subfigure{
\scalebox{0.19}[0.19]{\includegraphics{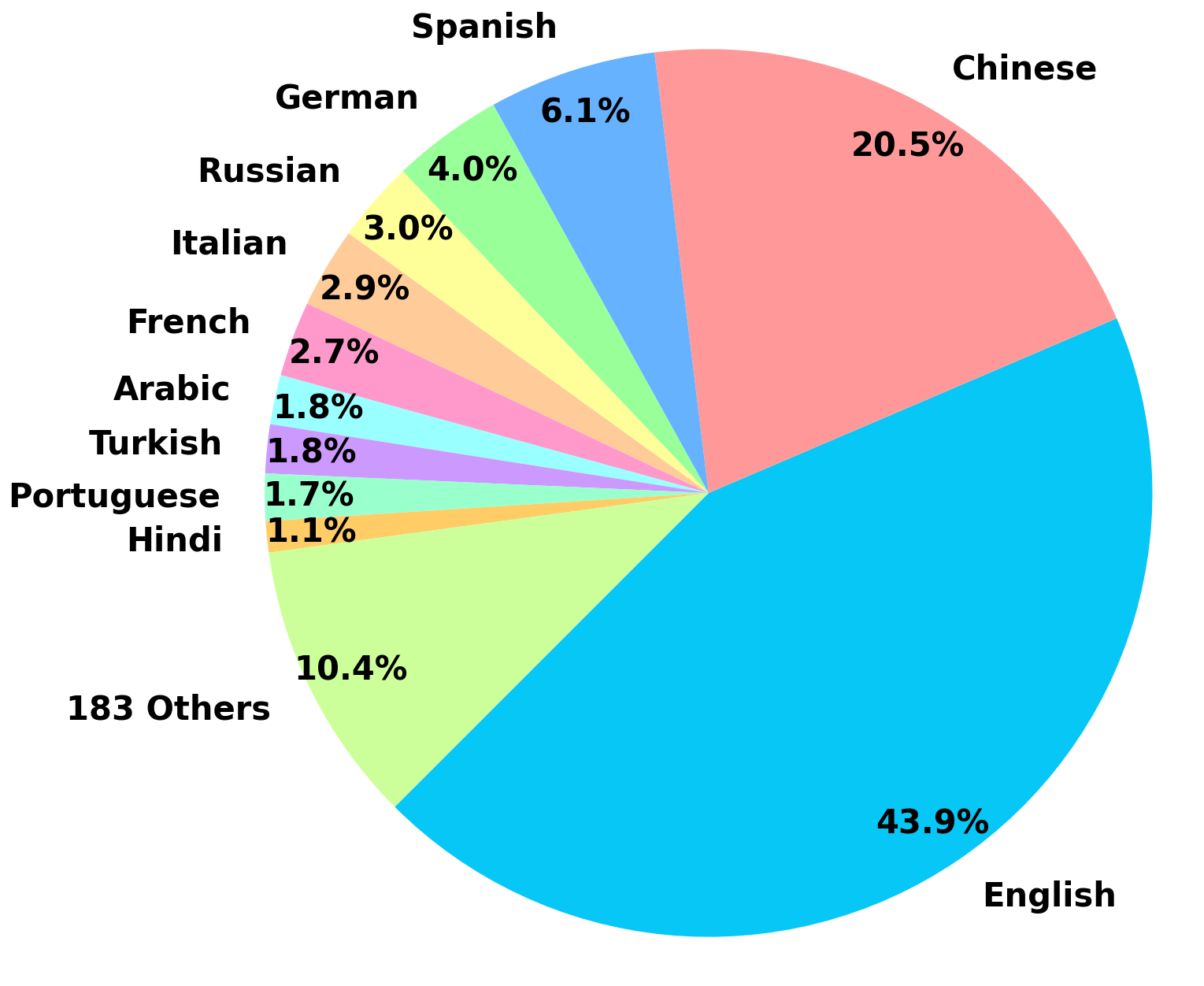}}
}
\subfigure{
\scalebox{0.19}[0.19]{\includegraphics{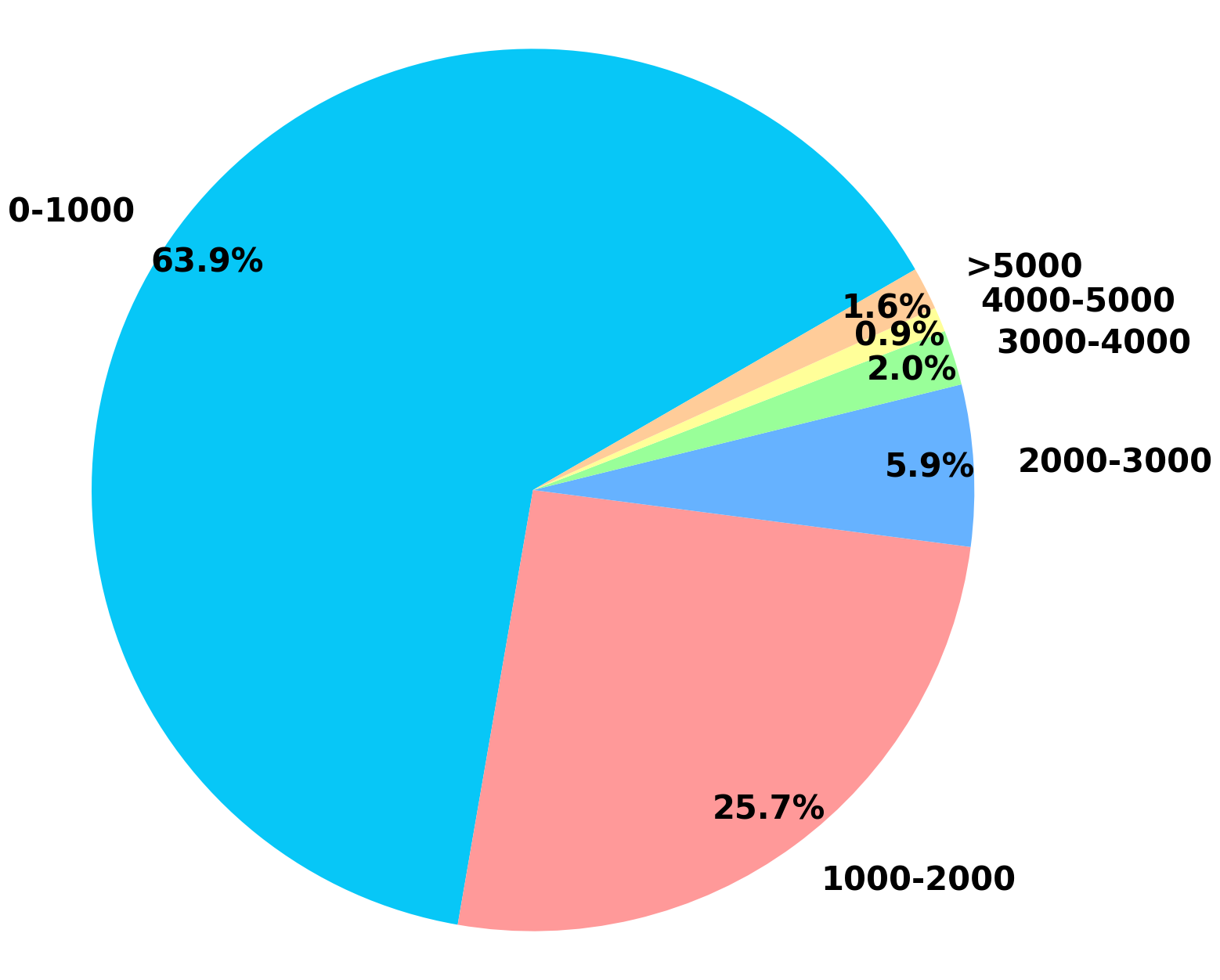}}
}
\caption{Language and sequence length distribution of unsupervised data}
\label{fig:unsupervised_data}
\end{figure*}

The language and length distribution (the number of tokens) of the unsupervised data are illustrated in Figure~\ref{fig:unsupervised_data}.

 We observed that for long texts (e.g., the news in cc-news), the initial sentences tend to be summarizing statements, and the model can rely solely on the information presented in these initial sentences to establish relevant relationships. To prevent the model from focusing solely on these starting sentences, we implemented a strategy of randomly shuffling the order of segments within entire texts. Specifically, we divided the text into three segments, shuffled their order randomly, and recombined them. This approach allows relevant text segments to appear randomly at any position within the long sequence. During training, we applied this operation to passages with a probability of 0.2\%.

\subsection{Synthetic Data}
\label{appendix:syn_data}
The prompt for GPT3.5 is ``You are a curious AI assistant, please generate one specific and valuable question based on the following text. The generated question should revolve around the core content of this text, and avoid using pronouns (e.g., "this"). Note that you should generate only one question, without including additional content:''.
The details of generated dataset are shown in Table~\ref{tab:bge_long}.


\section{Implementation Details}

\begin{table}[!t]
    \centering
    \setlength{\tabcolsep}{4pt}
    \begin{tabular}{lcc}
    \toprule
        \multirow{2}{*}{Length Range} & \multicolumn{2}{c}{Batch Size} \\
        \cline{2-3}
            & Unsupervised &  Fine-tuning \\
        \hline
        0-500       & 67,200 & 1,152 \\
        500-1000    & 54,720 & 768 \\
        1000-2000   & 37,248 & 480 \\
        2000-3000   & 27,648 & 432 \\
        3000-4000   & 21,504 & 336 \\
        4000-5000   & 17,280 & 336 \\
        5000-6000   & 15,072 & 288 \\
        6000-7000   & 12,288 & 240 \\
        7000-8192   & 9,984  & 192 \\
    \bottomrule
    \end{tabular}
    \caption{Detailed total batch size used in training for data with different sequence length ranges.}
    \label{tab:batch_size_dict}
\end{table}

\subsection{Experimental Hyperparameters} \label{appendix_sec:hyperparameters}

We adopt a further pre-trained XLM-RoBERTa\footnote{\scriptsize \href{https://huggingface.co/FacebookAI/xlm-roberta-large}{https://huggingface.co/FacebookAI/xlm-roberta-large}} as the foundational model. 
We extend the max position to 8192 and update the model via the RetroMAE~\cite{xiao-etal-2022-retromae} method.
The data comprises Pile~\cite{pile}, Wudao~\cite{yuan2021wudaocorpora}, and mC4~\cite{2019t5} datasets. We sampled a total of 184 million text samples from these sources, covering 105 languages. 
The maximum sequence length is 8192 and the learning rate is $7\times 10^{-5}$. 
The batch size is set to 32 and we accumulate the gradient over 16 steps.
 Pre-training is conducted on 32 A100(40GB) GPUs for 20,000 steps.

For the pre-training with the massive unsupervised data, the max length of query and passage is set to 512 and 8192, respectively. The learning rate is $5\times 10^{-5}$, the warmup ratio is 0.1 and the weight decay is 0.01. This training process takes 25,000 steps.
For training data with different sequence length ranges (e.g., 0-500, 500-1000, etc.), we use different batch sizes. The details are represented in Table~\ref{tab:batch_size_dict}. 
The second stage is conducted on 96 A800(80GB) GPUs.

In the fine-tuning stage, we sample 7 negatives for each query.
Refer to Table~\ref{tab:batch_size_dict} for the batch size. 
In the initial phase, we employed approximately 6000 steps to perform warm-up on dense embedding, sparse embedding and multi-vectors. Subsequently, we conducted unified training with self-knowledge distillation. These experiments were carried out on 24 A800(80GB) GPUs.

\subsection{MCLS Method}
\label{appendix:mcls}
The fine-tuning using long text can be constrained
due to the absence of long text data or computation
resources. In this situation, we propose a simple
but effective method: MCLS(Multiple CLS) to enhance the model’s ability without fine-tuning on
long text. The MCLS method aims to utilize multiple CLS tokens to jointly capture the semantics of long texts. Specifically, we insert a CLS token for every fixed number of tokens (in our experiments, we insert a ``[CLS]'' for each 256 tokens), and each CLS token can capture semantic information from its neighboring tokens. Ultimately, the final text embedding is obtained by averaging the last hidden states of all CLS tokens.

\subsection{Split-batch Method} \label{appendix_sec:efficient_batching}

\begin{algorithm}[h]
\caption{Pseudocode of split-batch.}
\label{alg:code}
\begin{lstlisting}[language=python]
# enable gradient-checkpointing
M3.gradient_checkpointing_enable()

embs = []
for batch_data in loader:
    # split the large batch into multiple sub-batch
    for sub_batch_data in batch_data:
        sub_emb = M3(sub_batch_data)
        # only collect the embs
        embs.append(sub_emb) 

# concatenate the outputs to get final embeddings
embs = cat(embs)
\end{lstlisting}
\end{algorithm}

Algorthm~\ref{alg:code} provides the pseudo-code of the split-batch strategy. 
For the current batch, we partition it into multiple smaller sub-batches. For each sub-batch we utilize the model to generate embeddings, discarding all intermediate activations via gradient checkpointing during the forward pass. Finally, we gather the encoded results from all sub-batch, and obtain the embeddings for the current batch. It is crucial to enable the gradient-checkpointing strategy; otherwise, the intermediate activations for each sub-batch will continuously accumulate, ultimately occupying the same amount of GPU memory as traditional methods.

In Table~\ref{tab:efficient_batching}, we investigate the impact of split-batch on batch size. It can be observed that, with the split-batch enabled, there is a significant increase in batch size. Simultaneously, the increase becomes more pronounced with longer text lengths, and in the case of a length of 8192, enabling split-batch results in a growth of batch size by over 20 times.

\begin{table}[h]
    \centering
    \setlength{\tabcolsep}{5pt}
    \begin{tabular}{cccc}
    \hline
    \multirow{2}{*}{Use Split-batch} & \multicolumn{3}{c}{Max Length} \\
    \cline{2-4}
    & 1024 & 4096 & 8192 \\
    \hline
    $\times$ & 262 & 25  & 6 \\
    $\surd$ & 855 & 258 & 130 \\
    \hline
    \end{tabular}
    \caption{Maximum batch size per device under different experimental settings.}
    \label{tab:efficient_batching}
\end{table}


\section{More Results} 

\subsection{Additional Resutls}
\label{appendix_sec:additional_results}
In this section, we present additional evaluation results on the MIRACL and MKQA benchmarks. As shown in Table~\ref{tab:miracl_recall_results} and 
 \ref{tab:mkqa_recall@20_results}, M3-Embedding outperforms all baselines on average. 

The detailed results of ablation studies of self-knowledge distillation and multi-stage training on the MIRACL dev set are shown in Table~\ref{tab:unify_ablation} and Table~\ref{tab:stage_ablation}.

\begin{figure}[ht]
    \centering
    \scalebox{0.3}[0.3]
    {\includegraphics{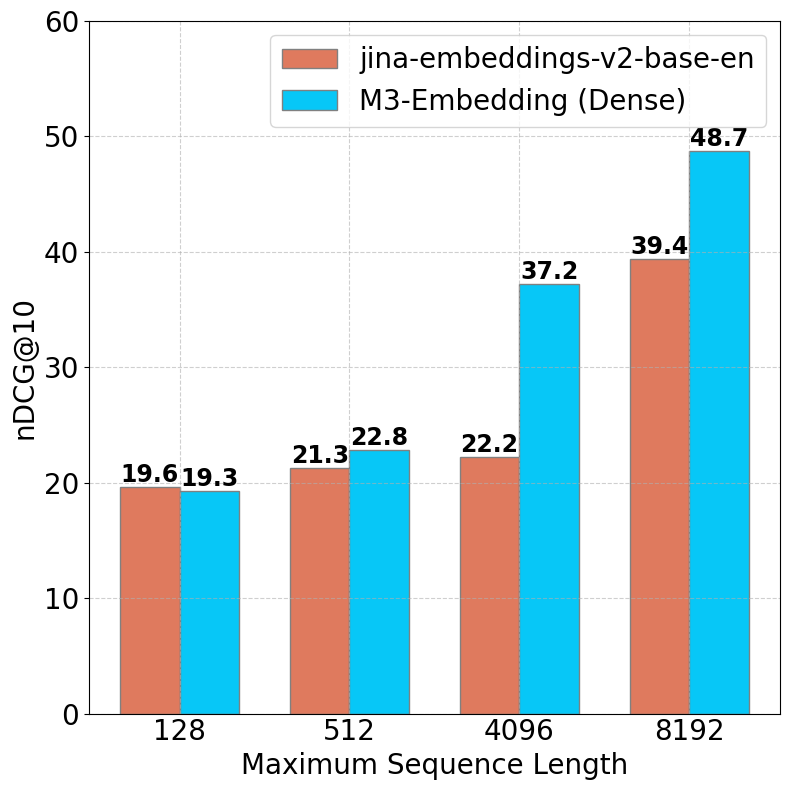}}
    \vspace{-5pt}
    \caption{NarrativeQA with variant sequence length.}
    \label{fig:narrativeqa_trend}
\end{figure}

\subsection{Different Tokenizer for BM25}
\label{appendix:tokenizer_bm25}

We investigate the impact of different tokenizers on the BM25 method, and the results are shown in Table~\ref{tab:bm25_tokenizer}. We can observe that:
\begin{itemize}
    \item Using the Analyzer from Lucene\footnote{\scriptsize \href{https://github.com/apache/lucene/tree/main/lucene/analysis/common/src/java/org/apache/lucene/analysis}{https://github.com/apache/lucene/tree/main/lucene/analysis/common/src/java/org/apache/lucene/analysis}} can significantly enhance the effectiveness of BM25. Lucene analyzer includes multiple steps typically including tokenization, stemming, stopword removal, etc, achieving better results than directly using the tokenzier of XLM-RoBERTa. Additionally, it's worth noting that the vocabulary size of the tokenizer from XLM-RoBERTa is limited, resulting in fewer unique tokens after encoding documents (for example, on the MLDR dataset, the tokenizer of XLM-RoBERTa produces 1056 unique terms per article, while Lucene's analyzer generates 1451 unique terms, which is over 37\% more and will increase retrieval latency).

    \item M3 outperforms BM25 models using the same tokenizer on all datasets, indicating that the learned weights are significantly better than the weights calculated by BM25.

    \item The sparse retrieval of M3 outperforms BM25 on MIRACL and MKQA datasets. In long document retrieval (MLDR), M3's sparse doesn't surpass BM25 but achieves competitive performance. This suggests that BM25 remains a highly competitive baseline model. Exploring tokenizers that perform better for sparse representation is a worthwhile topic for future research.
\end{itemize}

\begin{table}[!t]
    \centering
    \setlength{\tabcolsep}{1.5pt}
    \begin{tabular}{llccc}
    \toprule
        Method & Tokenizer & MIRACL & MKQA & MLDR \\
        \midrule
        BM25 & Analyzer & 38.5 & 40.9 & 64.1 \\
        BM25 & XLM-R &  31.9 & 39.9 & 53.6 \\
        M3(Sparse) & XLM-R & 53.9 & 45.3 & 62.2 \\
        M3(All) & XLM-R & 71.5 & 75.5 & 65.0 \\
        
    \bottomrule
    \end{tabular}
    \caption{Comparison with the BM25 methods using different tokenizers. }
    \label{tab:bm25_tokenizer}
\end{table}

\begin{table*}[!t]
    \centering
    \footnotesize
    \setlength{\tabcolsep}{2.1pt}
    \begin{tabular}{l|c|cccccccccccccccccc}
    \hline
    Model & Avg & ar & bn & en & es & fa & fi & fr & hi & id & ja & ko & ru & sw & te & th & zh & de & yo \\
    \hline
    \multicolumn{20}{l}{Baselines~(\textit{Prior Work})} \\
    \hline
    BM25 & 67.3 & 78.7 & 90.0 & 63.6 & 25.4 & 68.1 & 81.2 & 50.2 & 73.8 & 71.8 & 73.6 & 70.1 & 56.4 & 69.9 & 73.3 & 87.5 & 55.1 & 42.8 & 80.1 \\
    mDPR & 79.0 & 84.1 & 81.9 & 76.8 & 86.4 & 89.8 & 78.8 & 91.5 & 77.6 & 57.3 & 82.5 & 73.7 & 79.7 & 61.6 & 76.2 & 67.8 & 94.4 & 89.8 & 71.5  \\
    mContriever & 84.9 & 92.5 & 92.1 & 79.7 & 84.1 & 65.4 & 95.3 & 82.4 & 64.6 & 80.2 & 87.8 & 87.5 & 85.0 & 91.1 & 96.1 & 93.6 & 90.3 & 84.1 & 77.0 \\
    mE5$_{\mathrm{{\text{large}}}}$ & 94.1 & 97.3 & 98.2 & 87.6 & 89.1 & 92.9 & 98.1 & 90.6 & 93.9 & 87.9 & 97.1 & 93.4 & 95.5 & 96.7 & 99.2 & 98.9 & 93.3 & 90.7 & 93.1 \\
    E5$_{\mathrm{\text{mistral-7b}}}$ & 92.7 & 96.0 & 96.0 & 90.2 & 87.5 & 88.0 & 96.7 & 92.8 & 89.9 & 88.4 & 95.1 & 89.4 & 95.0 & 95.5 & 95.1 & 96.5 & 90.1 & 88.7 & 97.9 \\
    \hline
    \multicolumn{20}{l}{M3-Embedding~(\textit{Our Work})} \\
    \hline
    Dense   & 95.5 & 97.6 & 98.7 & 90.7 & 91.1 & 94.0 & 97.9 & 93.8 & 94.4 & 90.5 & 97.5 & 95.5 & 95.9 & 97.2 & \textbf{99.4} & 99.1 & 96.9 & 90.9 & 98.7 \\
    Sparse  & 85.6 & 92.0 & 96.7 & 81.5 & 72.1 & 87.0 & 91.5 & 73.3 & 87.1 & 84.8 & 92.4 & 91.7 & 76.9 & 85.1 & 98.1 & 95.2 & 72.9 & 69.1 & 92.9 \\
    Multi-vec  & 96.3 & 97.8 & \textbf{98.9} & 91.7 & 92.4 & 94.9 & 98.2 & \textbf{96.1} & 95.1 & 92.5 & \textbf{98.0} & 95.9 & 96.6 & 97.3 & \textbf{99.4} & \textbf{99.2} & 97.3 & \textbf{92.4} & 99.2 \\
    Dense+Sparse  & 96.2 & \textbf{98.0} & \textbf{98.9} & \textbf{92.4} & 92.5 & \textbf{95.6} & 98.3 & 94.6 & \textbf{95.6} & \textbf{92.6} & 97.5 & 95.6 & 96.6 & \textbf{97.4} & 99.1 & 99.0 & 96.8 & 91.0 & \textbf{100.0} \\
    All  & \textbf{96.4} & \textbf{98.0} & \textbf{98.9} & 92.1 & \textbf{92.9} & \textbf{95.6} & \textbf{98.4} & 95.6 & 95.2 & 92.5 & \textbf{98.0} & \textbf{96.0} & \textbf{96.7} & 97.2 & \textbf{99.4} & \textbf{99.2} & \textbf{97.6} & 92.3 & 99.2 \\
    \hline
    \end{tabular}
    \caption{Recall@100 on the dev set of the MIRACL dataset for multilingual retrieval in all 18 languages.}
    \label{tab:miracl_recall_results}
\end{table*}

\begin{table*}[!t]
    \centering
    \small
    \setlength{\tabcolsep}{3pt}
    \begin{tabular}{lcccccc|ccccc}
    \hline
    & \multicolumn{6}{c|}{Baselines~(\textit{Prior Work})} & \multicolumn{5}{c}{M3-Embedding~(\textit{Our Work})} \\
    \hline
    & BM25 & mDPR & mContriever & mE5$_{\mathrm{{\text{large}}}}$ & E5$_{\mathrm{\text{mistral-7b}}}$ & OpenAI-3 & Dense & Sparse & Multi-vec & Dense+Sparse & All \\
    \hline
    ar      & 13.4 & 33.8 & 43.8 & 59.7 & 47.6 & 55.1 & 61.9 & 19.5 & 62.6 & 61.9 & \textbf{63.0} \\
    da      & 36.2 & 55.7 & 63.3 & 71.7 & \textbf{72.3} & 67.6 & 71.2 & 45.1 & 71.7 & 71.3 & 72.0 \\
    de      & 23.3 & 53.2 & 60.2 & \textbf{71.2} & 70.8 & 67.6 & 69.8 & 33.2 & 69.6 & 70.2 & 70.4 \\
    es      & 29.8 & 55.4 & 62.3 & 70.8 & \textbf{71.6} & 68.0 & 69.8 & 40.3 & 70.3 & 70.2 & 70.7 \\
    fi      & 33.2 & 42.8 & 58.7 & 67.7 & 63.6 & 65.5 & 67.8 & 41.2 & 68.3 & 68.4 & \textbf{68.9} \\
    fr      & 30.3 & 56.5 & 62.6 & 69.5 & \textbf{72.7} & 68.2 & 69.6 & 43.2 & 70.1 & 70.1 & 70.8 \\
    he      & 16.1 & 34.0 & 50.5 & 61.4 & 32.4 & 46.3 & 63.4 & 24.5 & 64.4 & 63.5 & \textbf{64.6} \\
    hu      & 26.1 & 46.1 & 57.1 & 68.0 & \textbf{68.3} & 64.0 & 67.1 & 34.5 & 67.3 & 67.7 & 67.9 \\
    it      & 31.5 & 53.8 & 62.0 & 71.2 & \textbf{71.3} & 67.6 & 69.7 & 41.5 & 69.9 & 69.9 & 70.3 \\
    ja      & 14.5 & 46.3 & 50.7 & 63.1 & 57.6 & 64.2 & 67.0 & 23.3 & 67.8 & 67.1 & \textbf{67.9} \\
    km      & 20.7 & 20.6 & 18.7 & 18.3 & 23.3 & 25.7 & 58.5 & 24.4 & 59.2 & 58.9 & \textbf{59.5} \\
    ko      & 18.3 & 36.8 & 44.9 & 58.9 & 49.4 & 53.9 & 61.9 & 24.3 & 63.2 & 62.1 & \textbf{63.3} \\
    ms      & 42.3 & 53.8 & 63.7 & 70.2 & 71.1 & 66.1 & 71.6 & 52.5 & 72.1 & 71.8 & \textbf{72.3} \\
    nl      & 42.5 & 56.9 & 63.9 & \textbf{73.0} & 74.5 & 68.8 & 71.3 & 52.9 & 71.8 & 71.7 & 72.3 \\
    no      & 38.5 & 55.2 & 63.0 & 71.1 & 70.8 & 67.0 & 70.7 & 47.0 & 71.4 & 71.1 & \textbf{71.6} \\
    pl      & 28.7 & 50.4 & 60.9 & 70.5 & \textbf{71.5} & 66.1 & 69.4 & 36.4 & 70.0 & 69.9 & 70.4 \\
    pt      & 31.8 & 52.5 & 61.0 & 66.8 & \textbf{71.6} & 67.7 & 69.3 & 40.2 & 70.0 & 69.8 & 70.6 \\
    ru      & 21.8 & 49.8 & 57.9 & \textbf{70.6} & 68.7 & 65.1 & 69.4 & 29.2 & 70.0 & 69.4 & 70.0 \\
    sv      & 41.1 & 54.9 & 62.7 & 72.0 & \textbf{73.3} & 67.8 & 70.5 & 49.8 & 71.3 & 71.5 & 71.5 \\
    th      & 28.4 & 40.9 & 54.4 & 69.7 & 57.1 & 55.2 & 69.6 & 34.7 & 70.5 & 69.8 & \textbf{70.8} \\
    tr      & 33.5 & 45.5 & 59.9 & 67.3 & 65.5 & 64.9 & 68.2 & 40.9 & 69.0 & 69.1 & \textbf{69.6} \\
    vi      & 33.6 & 51.3 & 59.9 & 68.7 & 62.3 & 63.5 & 69.6 & 42.2 & 70.5 & 70.2 & \textbf{70.9} \\
    zh\_cn  & 19.4 & 50.1 & 55.9 & 44.3 & 61.2 & 62.7 & 66.4 & 26.9 & 66.7 & 66.6 & \textbf{67.3} \\
    zh\_hk  & 23.9 & 50.2 & 55.5 & 46.4 & 55.9 & 61.4 & 65.8 & 31.2 & 66.4 & 65.9 & \textbf{66.7} \\
    zh\_tw  & 22.5 & 50.6 & 55.2 & 45.9 & 56.5 & 61.6 & 64.8 & 29.8 & 65.3 & 64.9 & \textbf{65.6} \\
    \hline
    Avg     & 28.1 & 47.9 & 56.3 & 63.5 & 62.4 & 62.1 & 67.8 & 36.3 & 68.4 & 68.1 & \textbf{68.8} \\
    \hline
    \end{tabular}
    \caption{Recall@20 on MKQA dataset for cross-lingual retrieval in all 25 languages.}
    \label{tab:mkqa_recall@20_results}
\end{table*}

\begin{table*}[!t]
    \centering
    \small
    \setlength{\tabcolsep}{2.5pt}
    \begin{tabular}{l|c|cccccccccccccccccc}
    \hline
    Model & Avg & ar & bn & en & es & fa & fi & fr & hi & id & ja & ko & ru & sw & te & th & zh & de & yo \\
    \hline
    \multicolumn{20}{l}{M3-w.skd} \\
    \hline
    Dense & 69.2 & 78.4 & 80.0 & 56.9 & 56.1 & 60.9 & 78.6 & 58.3 & 59.5 & 56.1 & 72.8 & 69.9 & 70.1 & 78.7 & 86.2 & 82.6 & 62.7 & 56.7 & 81.8 \\
    Sparse & 53.9 & 67.1 & 68.9 & 43.8 & 38.6 & 45.1 & 65.4 & 35.3 & 48.2 & 48.9 & 56.1 & 61.5 & 44.5 & 57.9 & 79.1 & 70.9 & 36.1 & 32.5 & 70.0 \\
    Multi-vec & 70.5 & 79.6 & 81.0 & 59.3 & 57.8 & 62.0 & 80.1 & 59.4 & 61.5 & 58.3 & 74.5 & 71.2 & 71.2 & 79.1 & 87.9 & 83.0 & 63.7 & 58.0 & 82.4 \\
    \hline
    \multicolumn{20}{l}{M3-w.o.skd} \\
    \hline
    Dense & 68.7 & 78.0 & 79.1 & 56.4 & 55.4 & 60.3 & 78.3 & 58.2 & 59.0 & 55.1 & 72.4 & 68.8 & 69.5 & 77.8 & 85.8 & 82.5 & 63.0 & 56.0 & 80.6 \\
    Sparse & 36.7 & 48.2 & 51.9 & 24.3 & 20.3 & 26.0 & 48.6 & 16.8 & 30.1 & 32.0 & 33.0 & 43.1 & 27.2 & 45.2 & 63.6 & 52.2 & 22.6 & 16.5 & 59.2 \\
    Multi-vec & 69.3 & 78.7 & 80.2 & 57.6 & 56.7 & 60.5 & 79.0 & 58.4 & 59.3 & 57.5 & 74.0 & 70.3 & 70.2 & 78.6 & 86.9 & 82.1 & 61.9 & 56.7 & 78.2 \\
    \hline
    \end{tabular}
    \caption{Ablation study of self-knowledge distillation on the MIRACL dev set (nDCG@10).}
    \label{tab:unify_ablation}
\end{table*}

\begin{table*}[!t]
    \centering
    \small
    \setlength{\tabcolsep}{2.5pt}
    \begin{tabular}{l|c|cccccccccccccccccc}
    \hline
    Model & Avg & ar & bn & en & es & fa & fi & fr & hi & id & ja & ko & ru & sw & te & th & zh & de & yo \\
    \hline
    \multicolumn{20}{l}{Fine-tune} \\
    \hline
    Dense & 60.5 & 71.0 & 72.5 & 47.6 & 46.7 & 51.8 & 72.3 & 50.9 & 48.9 & 48.9 & 65.7 & 60.5 & 60.9 & 71.9 & 81.3 & 74.7 & 54.4 & 48.7 & 60.6 \\
    \hline
    \multicolumn{20}{l}{RetroMAE + Fine-tune} \\
    \hline
    Dense & 66.1 & 75.9 & 77.9 & 54.5 & 54.0 & 58.3 & 76.6 & 55.1 & 57.0 & 53.9 & 70.1 & 66.9 & 66.9 & 74.8 & 86.1 & 79.5 & 61.9 & 52.7 & 67.5 \\
    \hline
    \multicolumn{20}{l}{RetroMAE + Unsup + Fine-tune} \\
    \hline
    Dense & 69.2 & 78.4 & 80.0 & 56.9 & 56.1 & 60.9 & 78.6 & 58.3 & 59.5 & 56.1 & 72.8 & 69.9 & 70.1 & 78.7 & 86.2 & 82.6 & 62.7 & 56.7 & 81.8 \\
    \hline
    \end{tabular}
    \caption{Ablation study of multi-stage training on the MIRACL dev set (nDCG@10).}
    \label{tab:stage_ablation}
\end{table*}

\end{document}